\documentclass[conference]{IEEEtran}
\usepackage{cite}
\usepackage{amsmath,amssymb,amsfonts}
\usepackage{algorithmic}
\usepackage{graphicx}
\usepackage{textcomp}
\usepackage{xcolor}
\usepackage{array}
\usepackage{longtable}
\usepackage{graphicx} 
\usepackage{subfig}    
\usepackage{tabu}
\usepackage{longtable}[=v4.13]

\usepackage{background}
\usepackage{xcolor}
\usepackage{hyperref}

\def\BibTeX{{\rm B\kern-.05em{\sc i\kern-.025em b}\kern-.08em
    T\kern-.1667em\lower.7ex\hbox{E}\kern-.125emX}}

\backgroundsetup{
  scale=1,
  color=black,
  opacity=1,
  angle=0,
  position=current page.south,
  vshift=10pt,
  contents={\textcolor{red}{© 2024 IEEE. For personal use only. Published version under doi:  \href{https://doi.org/10.1109/AICCSA63423.2024.10912544}{10.1109/AICCSA63423.2024.10912544}. }}
}

\begin{document}

\title{An Efficient Deep Learning-Based Approach to Automating Invoice Document Validation}

\author{\IEEEauthorblockN{Aziz Amari}
\IEEEauthorblockA{\textit{National Institute of Applied} \\
\textit{Sciences and Technology (INSAT)} \\
\textit{University of Carthage} \\
Tunis, Tunisia \\
azizamari@ieee.org}
\and
\IEEEauthorblockN{{Mariem Makni}
\IEEEauthorblockA{\textit{National Institute of Applied} \\
\textit{Sciences and Technology (INSAT)} \\
\textit{University of Carthage} \\
Tunis, Tunisia \\
mariem.makni@insat.ucar.tn}
\and
\IEEEauthorblockN{Wissal Fnaich}
\IEEEauthorblockA{\textit{National Institute of Applied} \\
\textit{Sciences and Technology (INSAT)} \\
\textit{University of Carthage} \\
Tunis, Tunisia \\
wissal.fnaich@insat.ucar.tn}
\and
\IEEEauthorblockN{Akram Lahmar}
\IEEEauthorblockA{\textit{National Institute of Applied} \\
\textit{Sciences and Technology (INSAT)} \\
\textit{University of Carthage} \\
Tunis, Tunisia \\
akram.lahmar@insat.ucar.tn}
\and
\IEEEauthorblockN{Fedi Koubaa}
\IEEEauthorblockA{\textit{National Institute of Applied} \\
\textit{Sciences and Technology (INSAT)} \\
\textit{University of Carthage} \\
Tunis, Tunisia \\
fedi.koubaa@insat.ucar.tn}
\and
\IEEEauthorblockN{Oumayma Charrad}
\IEEEauthorblockA{\textit{National Institute of Applied} \\
\textit{Sciences and Technology (INSAT)} \\
\textit{University of Carthage} \\
Tunis, Tunisia \\
oumayma.charrad@ieee.org}
\and
\IEEEauthorblockN{Mohamed Ali Zormati}
\IEEEauthorblockA{\textit{National Institute of Applied} \\
\textit{Sciences and Technology (INSAT)} \\
\textit{University of Carthage} \\
Tunis, Tunisia \\
zormati@ieee.org}}
\and
\IEEEauthorblockN{Rabaa Youssef Douss}
\IEEEauthorblockA{\textit{National Institute of Applied} \\
\textit{Sciences and Technology (INSAT)} \\
\textit{COSIM Lab. LR11TIC01} \\
\textit{University of Carthage} \\
Tunis, Tunisia \\
rabaa.youssef@insat.ucar.tn}
}

\maketitle
\begin{abstract}
In large organizations, the number of financial transactions can grow rapidly, driving the need for fast and accurate multi-criteria invoice validation. Manual processing remains error-prone and time-consuming, while current automated solutions are limited by their inability to support a variety of constraints, such as documents that are partially handwritten or photographed with a mobile phone. In this paper, we propose to automate the validation of machine written invoices using document layout analysis and object detection techniques based on recent deep learning (DL) models. We introduce a novel dataset consisting of manually annotated real-world invoices and a multi-criteria validation process. We fine-tune and benchmark the most relevant DL models on our dataset. Experimental results show the effectiveness of the proposed pipeline and selected DL models in terms of achieving fast and accurate validation of invoices.
\end{abstract}


\section{Introduction}
In large organizations, it is a common case to manually process a steady stream of invoice documents, for example, to reimburse staff expenses. Invoice processing for validation purposes is quite challenging. In fact, proof of authenticity mainly relies on the presence of a handwritten signature and a stamp \cite{b0}, and the verification of key elements of the document (e.g., document type such as receipt and invoice, client name, date of issue, etc.). It is therefore a necessity to consider automated processing, which requires handling the variability of document layout and quality \cite{b1}, in order to guarantee reliable invoice validation.

To enable efficient understanding and processing of invoice documents, it is appropriate to consider deep learning (DL) techniques, which have revolutionized the accuracy and efficiency of both text recognition and object detection and have shown remarkable results in various applications \cite{bx1, bx2}. For the text recognition needed for invoice validation, many recent methods, such as LiLT \cite{b11}, LayoutLMv2 \cite{b12}, and LayoutLMv3 \cite{b10} are at the forefront, allowing the understanding of complex structured documents, such as invoices. These attention-based models include embedded optical character recognition (OCR) to enhance document understanding. Regarding object detection, a wide range of deep architectures are proposed in the literature, including FCOS \cite{b15}, Faster R-CNN \cite{x2}, RetinaNet \cite{b16}, and SSD \cite{b17}. These DL models can potentially meet the need of identifying and localizing stamp and signature, which is an underlying task of invoice validation.

When considering DL models for invoice validation, and since these models are not specifically trained on this particular task, it is relevant to consider transfer learning, where models pre-trained on large datasets can be fine-tuned on a smaller dataset specific to invoices. Recent works applied transfer learning on financial document datasets \cite{b18,b19,b20}, which improved model performance with limited data. However, after reviewing these datasets, we note that some of them are of receipt type \cite{b18,b19}, others consider a variety of forms \cite{b20,b21}, which differ slightly from a regular machine written invoice. In \cite{b22}, the authors propose a large dataset of 10000 synthetic invoices, specifically tailored for document analysis and understanding. However, these documents do not always reflect real-world scenarios where invoices are often photographed with mobile phones, introducing significant distortions. We also notice that many of the datasets lack the presence of stamp and signature components.

In order to fill this gap and to effectively apply DL techniques, we propose a novel, fully annotated dataset composed of real machine written invoice documents with various imperfections, such as blurriness, class imbalance, low-quality phone-shot images, and varying formats. In our case study, we consider a machine written invoice document to be valid only if it contains a stamp, a signature, and certain selected fields. After collecting and annotating the dataset, we select the most relevant DL-based approaches for implementation and benchmarking. We progressively evaluate and improve the results until obtaining a robust pipeline that maximizes accuracy and reliability, while maintaining reasonable performance.

The remainder of this paper is structured as follows. In Section \ref{sec:mat_meth}, we present the material considered and the methods proposed. The results are reported and discussed in Section \ref{sec:result_disc}, while Section \ref{sec:concl} concludes the paper and discusses the future perspectives.

\section{Materials and Methods\label{sec:mat_meth}}
\begin{figure}[h]
        \centering
        \includegraphics[width=0.5\textwidth]{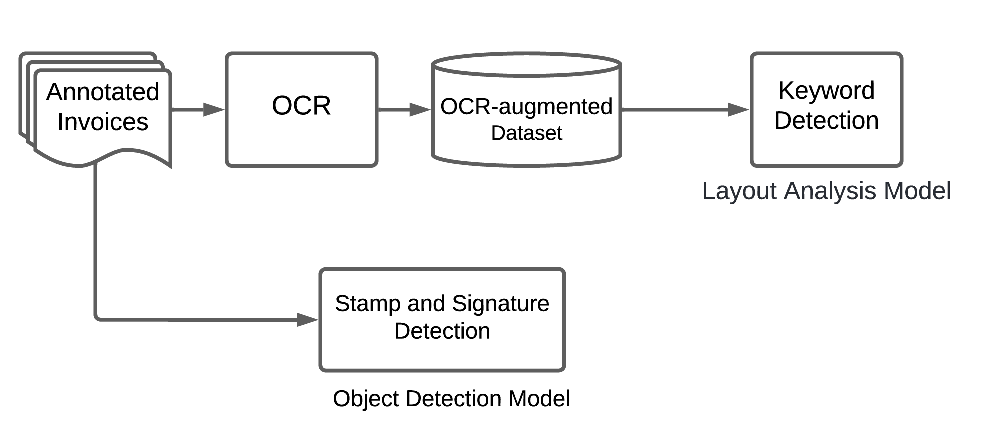}
        \caption{Adopted training pipeline.}
        \label{fig2}
    \end{figure}

In this section, we present the materials, detailing the data collection and processing pipeline, and then describe the methods, including the selection of the most recent and relevant models for implementation. Since invoice validation consists of two main tasks, we divide our work into two independent steps: keyword detection and object detection. The adopted training pipeline is shown in Figure \ref{fig2}, which shows that keyword detection is preceded by an OCR phase, which allows for an OCR-augmented dataset. For object detection, we aim to verify the presence of signatures and stamps to ensure the authenticity of financial documents.

\subsection{Materials}
We outline the dataset used for our DL pipeline by first describing the data annotation process, followed by data augmentation and preparation.

\subsubsection{Data annotation}~

We collected 381 real financial documents, of which 84\% (320 documents) are invoices. Among these, 240 are machine-written. We manually annotated the documents with bounding boxes around the necessary elements to establish the validity of the invoices, as shown in Table \ref{tab1}. All classes are represented as bounding boxes, except for a boolean attribute indicating whether handwriting is present.

\begin{table}[h]
\caption{Annotated dataset description.}
\label{tab1}
\centering
\begin{tabular}{ll}
\hline
\textbf{Class}  &\textbf{Description}                          \\ \hline
Title          & Document type (e.g., invoice, receipt, etc.). \\ \hline
Client         & Client name.                                  \\ \hline
Stamp          & Stamp for document authenticity.              \\ \hline
Signature      & Signature for document authenticity.          \\ \hline
Date           & Issue date, different formats.                \\ \hline
Total          & Total label.                                  \\ \hline
Total Value    & Total numeric value.                          \\ \hline
Handwritten    & True if handwritten (not machine-written).    \\ \hline
\end{tabular}
\end{table}

\begin{figure}[h]
        \centering
        \includegraphics[width=0.4\textwidth]{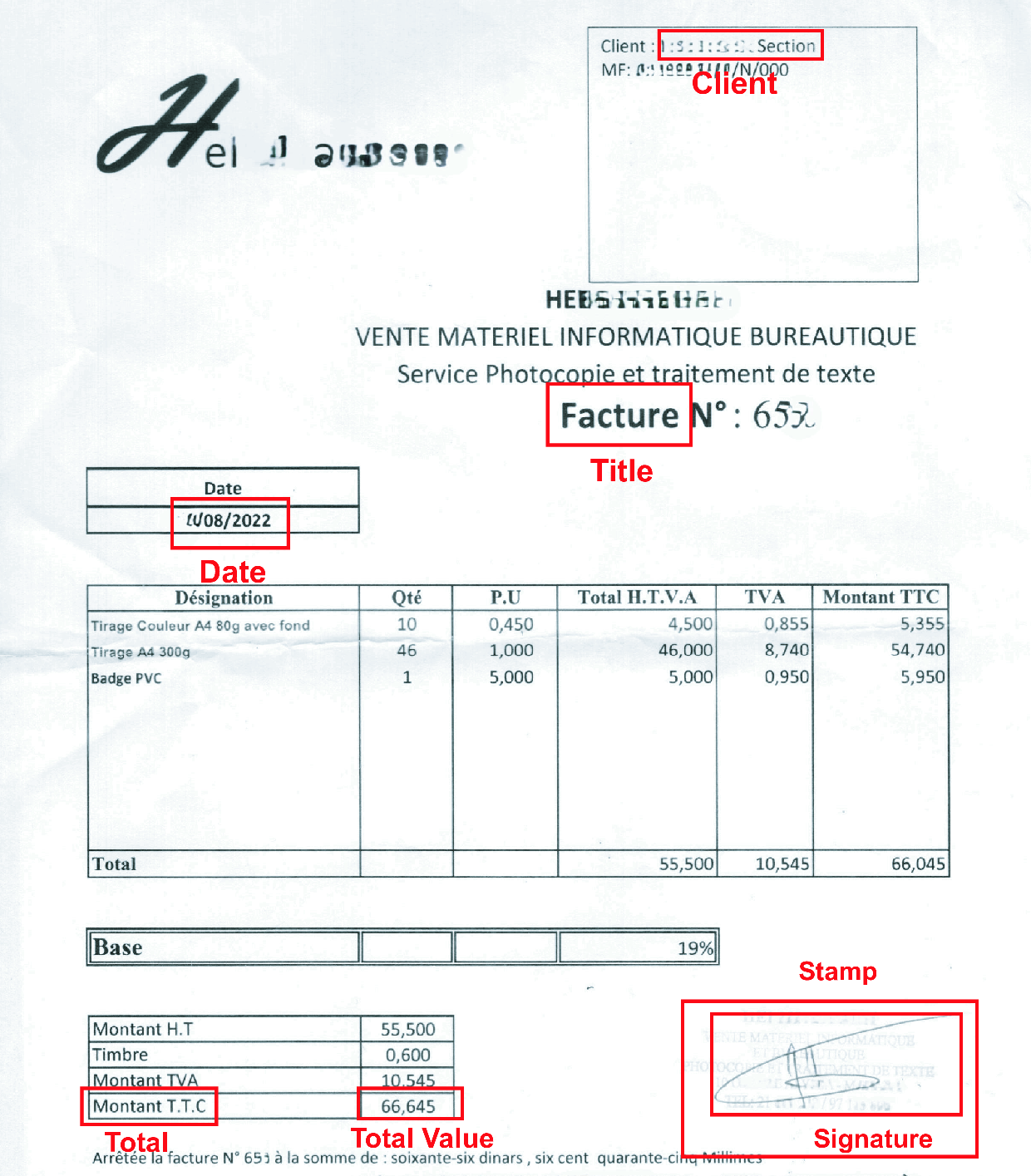}
        \caption{Example of an annotated document.}
        \label{fig3}
    \end{figure}

An example of an annotated document is shown in Figure \ref{fig3}. For confidentiality reasons, some parts of the financial documents have been blurred.

\begin{figure}[h]
        \centering
        \includegraphics[width=0.35\textwidth]{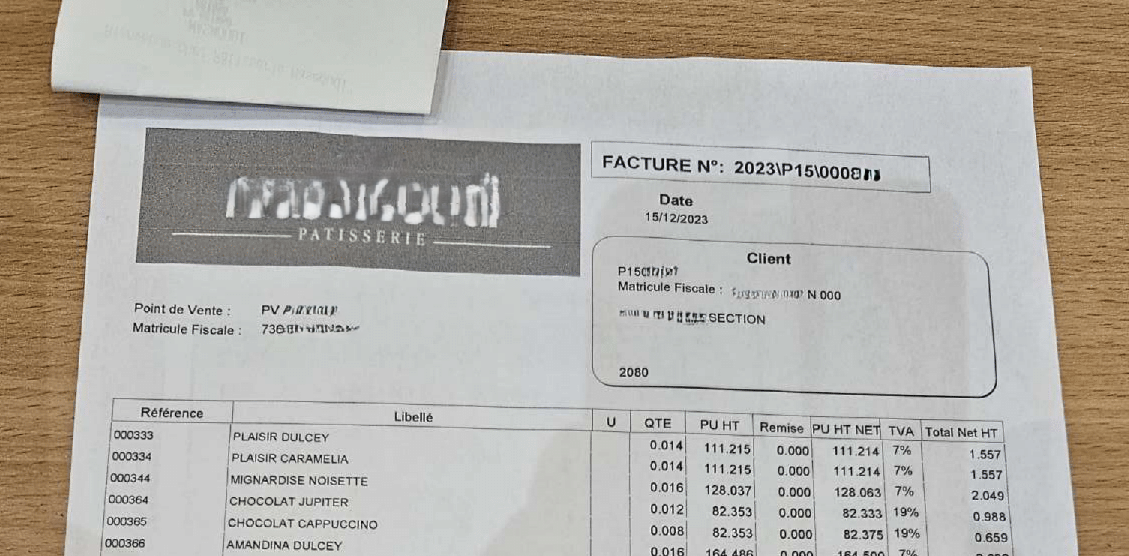}
        \caption{Example of a distorted document.}
        \label{fig1}
    \end{figure}

\begin{figure}[h]
        \centering
        \includegraphics[width=0.33\textwidth]{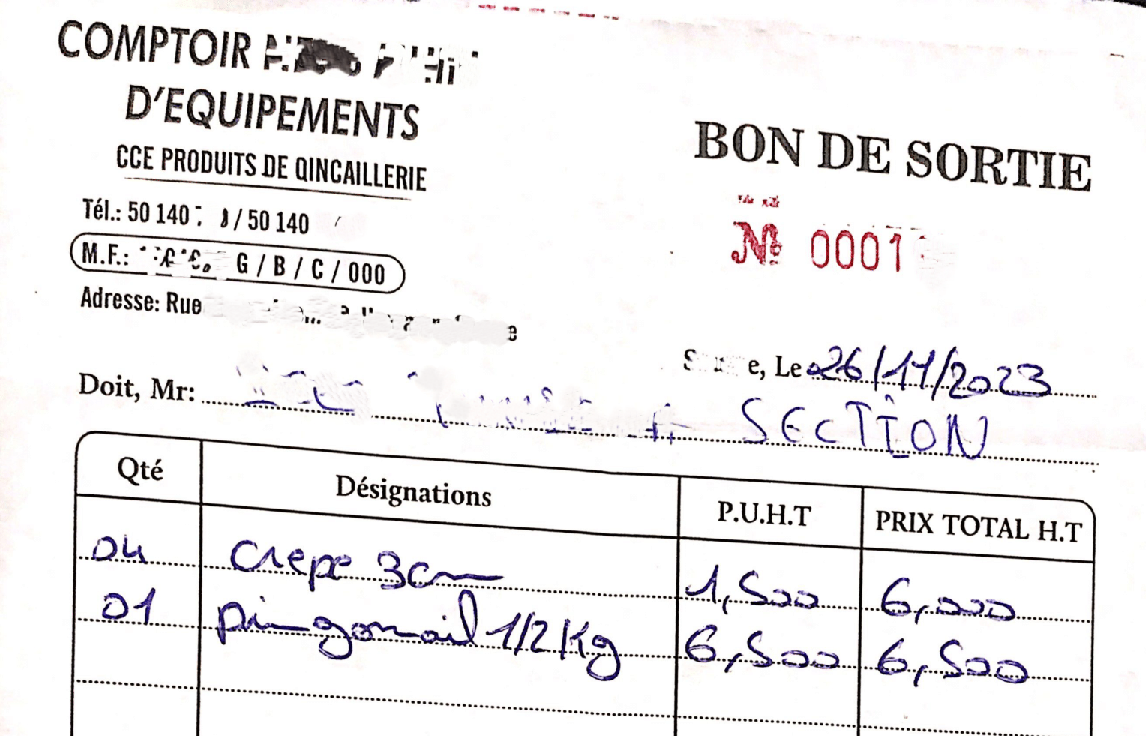}
        \caption{Example of a skewed document.}
        \label{fig4}
    \end{figure}

Our dataset includes various real invoices; many are distorted due to mobile phone capture as depicted in Figure \ref{fig1}, and some are handwritten, potentially appearing skewed during capture, as illustrated by Figure \ref{fig4}.

This work focuses on the processing of digital invoices, since  handwritten invoices are sparse in the current dataset, and present additional challenges for augmentation and processing.

\subsubsection{Data augmentation and preparation}~

In order to improve the ability of the DL models to generalize, data augmentation and preparation techniques are required due to the limited size of collected dataset. Given that the adopted workflow is divided into two main tasks, namely keyword detection and object detection, each task is subject to specific and adapted augmentation techniques.

For the keyword detection task, the challenges encountered by a DL model include the presence of unclear characters and image distortions such as rotation, shear, or scale transformations. Therefore, we choose to augment the dataset using the following techniques:
\begin{itemize} 
\item Median blur and color jitter: improves clarity by reducing noise and simulating various lighting conditions, particularly in documents with complex backgrounds.

\item Padding: prevents critical text from being cut off during data preparation, with a probability set to 1.

\item Rotation: is limited to 5 degrees using inter-cubic interpolation to minimize data loss. In order to preserve information and standardize image dimensions, an additional step ensures no cropping occurs closer than 20 pixels from any bounding box after rotation.

\item Resize: is performed using bi-linear sampling to maintain consistent image dimensions, supporting uniform model performance. 
\end{itemize}

For the stamp and signature detection task, our goal is to present these objects under varying conditions. The following outlines the augmentation techniques we apply:

\begin{itemize}
    \item Rotation: is performed with a maximum of 10 degrees to simulate the common skew in scanned documents. The rotation transformation is applied with a probability of 0.5 to ensure data diversity.
    
    \item Color jitter: is performed with a probability of 0.4 to account for color variation and lighting conditions, which  enhances the model's resilience in detecting signatures and stamps across different documents.

    \item Multiplicative noise: is added on a per-channel basis with a multiplier ranging from 0.5 to 1.0, applied with a probability of 0.3. This noise simulates digital scanning artifacts and helps train the model to be less sensitive to such variations.
\end{itemize}

\begin{figure}[h]
    \centering
    \includegraphics[width=0.42\textwidth]{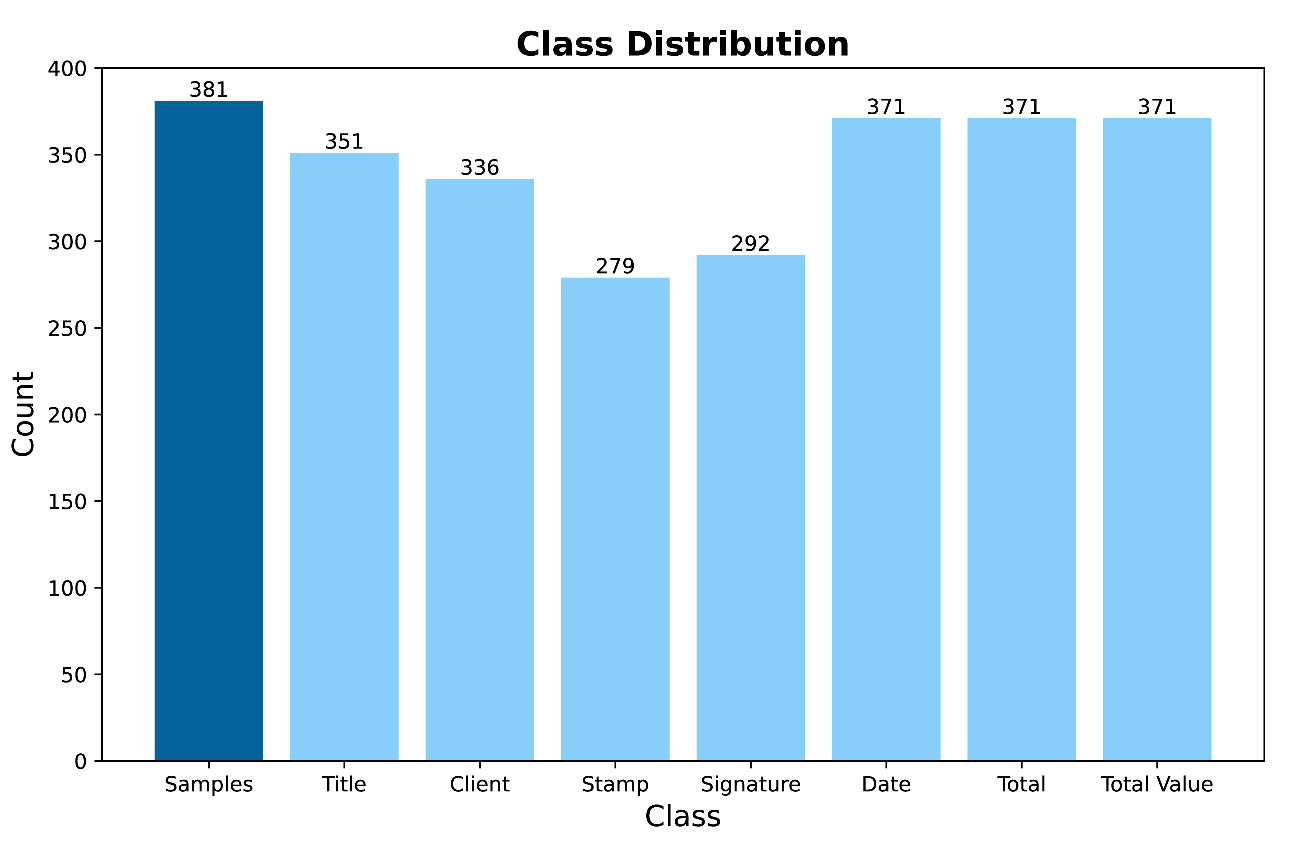}
    \caption{Final dataset class distribution.}
    \label{fig5}
\end{figure}

Figure \ref{fig5} illustrates the distribution of the final dataset after data augmentation. It is important to note that the class distribution remains unchanged following the augmentation process, as the applied operations only affect the number of instances per class.

\subsection{Methods}
This subsection outlines the deep learning models we plan to train and evaluate for both keyword detection and stamp and signature detection tasks. 

\subsubsection{Keyword detection}~

We perform keyword detection to identify significant parts of the document (e.g., customer name, document type, date, etc.). To achieve this, we use (i) an OCR step followed by (ii) a document layout analysis step. This approach is necessary for attention-based models that incorporate embedded OCR to enhance document understanding. Below, we present the selected methods for both steps and provide a comparison to determine the most effective combination of OCR and document layout analysis methods for keyword detection.

An OCR step for character extraction from input images, utilizing robust modules, is essential for the document layout analysis task. Choosing the appropriate OCR model requires comparing the most relevant OCR tools. The most relevant OCR models in the literature are tested on the FUNSD dataset \cite{b20}, which stands for Form Understanding in Noisy Scanned Documents. To evaluate the similarity between the extracted and the ground truth text in the dataset, we consider two metrics from the original papers: accuracy and similarity based on \textit{Levenshtein} distance \cite{x7}, which is particularly suited for measuring the distance between characters in a word.

\begin{table}[h]
\caption{Comparison of OCR models on FUNSD \cite{b20}.}
\label{tab2}
\centering
\begin{tabular}{lll}
\hline
\textbf{Model} & \textbf{Accuracy} & \textbf{Similarity}  \\ \hline
EasyOCR & 0.284 & 0.719 \\ \hline
Tesseract \cite{b18} & 0.306 & 0.569 \\ \hline
Amazon Textract & 0.345 & 0.513 \\ \hline
Google OCR & 0.576 & 0.892 \\ \hline
Microsoft Azure OCR & \textbf{0.633} & \textbf{0.915} \\ \hline
\end{tabular}
\end{table}

Table \ref{tab2} summarizes the results of the most relevant OCR models according to original papers, showing that Microsoft Azure OCR outperforms all methods and is the best cloud-based solution. Therefore, Microsoft Azure OCR is selected as our primary method. Additionally, we consider using Tesseract during implementation to explore the potential of a local model and to confirm or refute our decision.

After the OCR step, we use document layout analysis models to automate the extraction, categorization, and interpretation of information from the invoice documents. Recently, attention-based architectures have become commonly used, since attention mechanisms allow models to focus on the most relevant parts of the input data. This improves their ability to handle complex documents. Three relevant models are considered in this work. 

First, LayoutLMv2Large \cite{b12} combines textual and visual data to interpret complex document structures. The model employs a multimodal transformer architecture with the following pre-training objectives: masked visual-language modeling to enhance context understanding, text-image alignment for better simultaneous processing of visual and textual data, and text-image matching to ensure that the model can distinguish between matched and unmatched text-image pairs. This method uses a cross-entropy loss function for masked language modeling tasks, optimizing accuracy and relevancy of predictions.

Second, we consider LayoutLMv3Large \cite{b10}, which also combines textual and visual data. It leverages a multimodal transformer architecture with several pre-training objectives: masked language modeling to enhance the understanding of textual context, masked image modeling to learn document layouts, and word-patch alignment to improve simultaneous processing of visual and textual data. As for LayoutLMv2Large \cite{b12}, this mùodel uses a cross-entropy loss function for masked language modeling tasks.

Third, LiLT \cite{b11} is also a relevant model that employs a dual-stream Transformer architecture, processing textual and layout information in parallel by integrating through a bi-directional attention complementation mechanism. It is pre-trained using: masked visual-language modeling to enhance language comprehension while preserving layout awareness, and key point location and cross-modal alignment identification to understand and align textual and visual elements within documents. This method also uses cross-entropy loss for masked visual-language modeling to handle multi-class classification problems effectively.

\begin{table}[h]
\caption{Comparison of document layout analysis models on RVL-CDIP \cite{b9}.}
\label{tab3}
\centering
\begin{tabular}{lll}
\hline
\textbf{Model} & \textbf{Accuracy} & \textbf{Parameters}  \\ \hline
LayoutLMv2Large \cite{b12} & 0.9564 & 425.9 millions \\ \hline
LayoutLMv3Large \cite{b10} & \textbf{0.9593} & 356.0 millions \\ \hline
LiLT \cite{b11} & 0.9568 & \textbf{130.0 millions} \\ \hline

\end{tabular}
\end{table}

In Table \ref{tab3}, we summarize the accuracy results of the three relevant document layout analysis models as evaluated on the RVL-CDIP dataset \cite{b9} according to original papers. We also include parameter count for the three models. From these results, we can deduce that LayoutLMv3Large \cite{b10} and LayoutLMv2Large \cite{b12} show high accuracy rates, making them strong candidates. LiLT \cite{b11} also performs well, though it has fewer parameters compared to other models, making it a good candidate when reduced inference time is needed.

\subsubsection{Signature and stamp detection}~

Object detection models locate instances of objects in images by estimating bounding boxes. In our case, we need to detect the presence of stamps and signatures to guarantee the authenticity of documents. We select four promising models for this purpose.

First, RetinaNet \cite{b16}, which is renowned for its approach in single-stage object detection, significantly enhances detection capabilities through feature pyramid networks and focal loss. Focal loss reduces the loss contribution from easy examples and increases the importance of correcting misclassified examples.

Second, the Fully Convolutional One-Stage (FCOS) model \cite{b15} operates on a per-pixel prediction basis, eliminating the need for anchor boxes, simplifying the detection pipeline, and reducing computational complexity. FCOS introduces several novel components in its loss function to handle object detection challenges effectively: classification loss, which is a binary cross-entropy loss that classifies whether each location is the center of an object, center-ness loss, which is a novel loss that down-weights the contribution of low-quality detections during training, focusing more on the center regions of objects, and regression loss, which is a bounding box regression loss using IoU-loss that measures the accuracy of the predicted boxes against the ground truth.

Third, Faster R-CNN \cite{x2} is renowned for its two-stage object detection approach, which significantly improves both speed and accuracy. The loss function in Faster R-CNN is a multi-task loss that combines classification loss (a softmax loss that evaluates the predicted class probabilities against the ground truth class labels) and bounding box regression loss (a smooth L1 loss that measures the difference between the predicted bounding box coordinates and the ground truth coordinates).

Finally, Single Shot MultiBox Detector (SSD) \cite{b17} is a single-stage object detection algorithm known for its balance between speed and accuracy. The loss function in SSD, as for Faster RCNN, is a combination of classification loss and bounding box regression loss.

\begin{table}[h]
\caption{Comparison of object detection models on COCO \cite{b14}.}
\label{tab4}
\centering
\begin{tabular}{lll}
\hline
\textbf{Model} & \textbf{mAP} & \textbf{Parameters}  \\ \hline
RetinaNet \cite{b16} & \textbf{41.5} & 38.2 millions \\ \hline
FCOS \cite{b15} & 39.2 & \textbf{32.3 millions} \\ \hline
Faster R-CNN \cite{x2} & 37 & 41.8 millions \\ \hline
SSD \cite{b17} & 25.1 & 35.6 millions \\ \hline

\end{tabular}
\end{table}

We present a comparison of the models using the number of parameters and their mean average precision (mAP) score on the Microsoft COCO dataset \cite{b14}. The results are summarized in Table \ref{tab4}. Among these models, RetinaNet \cite{b16} has the highest mAP, indicating superior performance in detecting and classifying objects within documents. We move forward with fine-tuning and testing the presented methods for stamp and signature detection.

Based on the comparisons made for keyword detection and stamp and signature detection, we summarize our choices as follows.
We select Microsoft Azure OCR and Tesseract for OCR step, based on their high accuracy and similarity scores. Regarding document layout analysis, we move forward with LayoutLMv3, LayoutLMv2, and LiLT, to be mainly evaluated using the F1-score. It is worth noting that we consider standard models (not large ones, i.e., LayoutLMv3Large and LayoutLMv2Large) for their lower resource comsumption and better fit for our data. For object detection, we benchmark RetinaNet, FCOS, Faster R-CNN, and SSD, using the mAP metric. These models are chosen based on their performance on benchmark datasets and their suitability for our specific tasks for invoice document validation.

\section{Results and Discussion\label{sec:result_disc}}

In this section, we present the results of the benchmarking performed on the selected models after fine-tuning them on our data. We also examine the challenge of class imbalance within the OCR-augmented dataset and its implications. In our experimental evaluation, we focus on precision, for a highly reliable invoice validation pipeline, with consideration of limited computational resources. The implementation of the models is performed using hardware consisting of an NVIDIA GeForce RTX 3050 GPU and the PyTorch framework. We first present and discuss results for keyword detection, then the results for signature and stamp detection.

\subsection{Evaluation of selected OCR models}

To ensure keyword detection reliability, we evaluate the two selected OCR methods: Microsoft Azure OCR and Tesseract. Since our dataset was not fully annotated for the OCR step, we propose to use these two OCR models in inference mode in order to produce annotations. Subsequently, we fine-tune the LayoutLMv3 model using these annotations and evaluate which of the two OCRs performs better with the fixed layout document analysis model. The training of LayoutLMv3 is performed under identical conditions, maintaining the same number of epochs and other hyperparameters.

\begin{table}[h]
\caption{Comparison of fine-tuning LayoutLMv3 using Tesseract versus Microsoft Azure OCR.}
\label{tab5}
\centering
\begin{tabular}{lllll}
\hline
\textbf{Method} & \textbf{Validation Loss} & \textbf{F1-Score} & \textbf{Recall} & \textbf{Precision} \\ \hline
Tesseract \cite{b18} & \textbf{$1e^{-2}$} & \textbf{0.81} & \textbf{0.77} & \textbf{0.87} \\ \hline
Azure OCR & $3.6e^{-2}$ & 0.56 & 0.48 & 0.68 \\ \hline
\end{tabular}
\end{table}

As shown in Table \ref{tab5}, Tesseract \cite{b18} demonstrates higher effectiveness with an F1-score of 0.81, indicating robust detection capabilities. In contrast, Microsoft Azure OCR’s results are more conservative, with an F1-score of 0.56 and a precision of 0.68. However, the variation in the dataset size handled by each OCR tool is substantial, as depicted in Table \ref{tab:class_imbalance}. Azure OCR’s ability to extract more instances on average provides a significant advantage, allowing the model to more accurately distinguish between irrelevant other (denoted as 'O') labels and targeted fields.

\begin{table}[h]
\caption{Comparison of class instance counts.}
\label{tab:class_imbalance}
\centering
\begin{tabular}{lllllll}
\hline
\textbf{Class} & \textbf{O} & \textbf{Title} & \textbf{Date} & \textbf{Client} & \textbf{Total} & \textbf{Total Value} \\ \hline
Azure OCR & 31715 & 217 & 242 & 553 & 315 & 221 \\ \hline
Tesseract & 6272 & 189 & 201 & 161 & 198 & 192 \\ \hline
\end{tabular}
\end{table}

\begin{figure}[h]
    \centering
\includegraphics[width=0.5\linewidth]{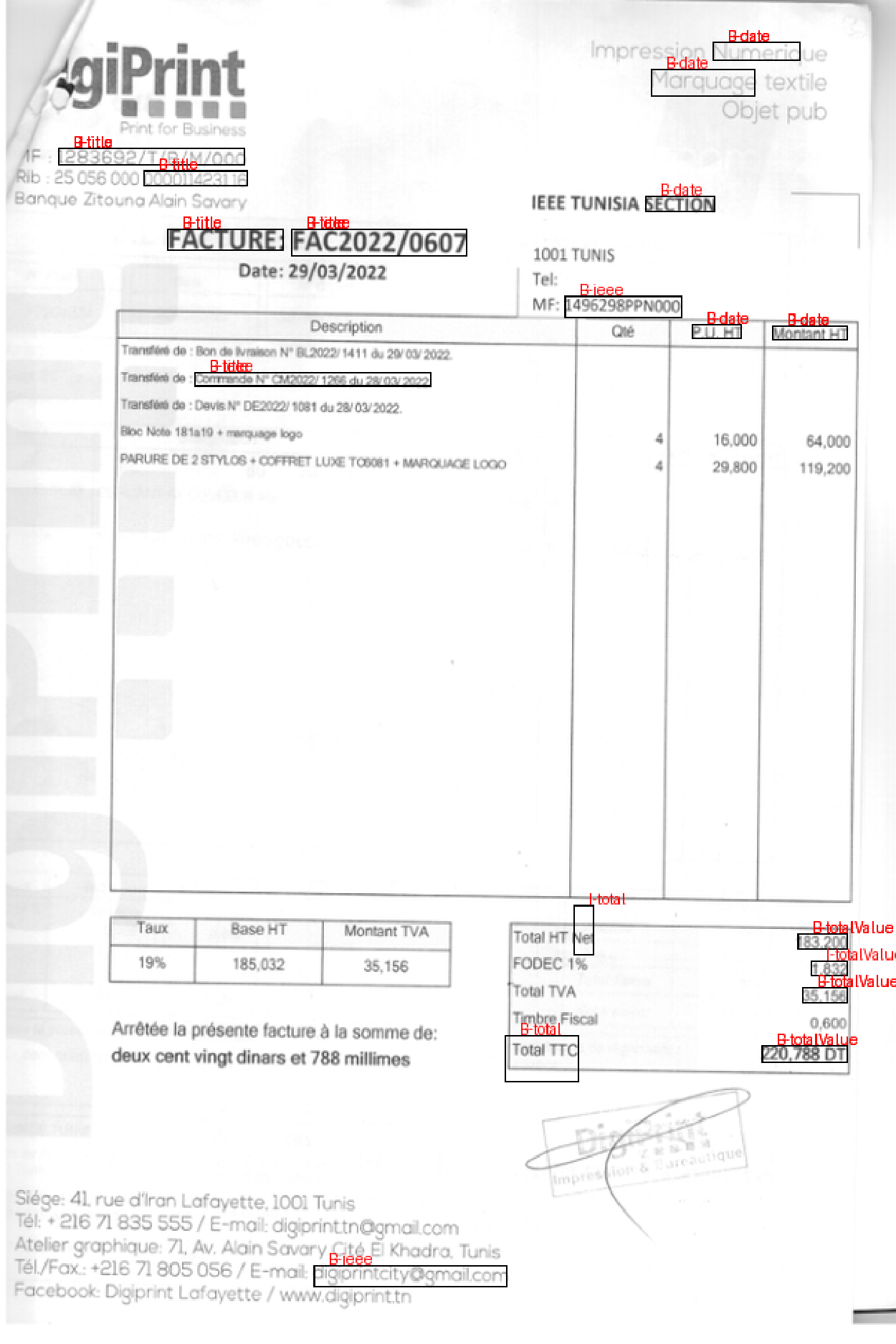}
    \caption{LayoutLMv3 output using Tesseract.}
    \label{fig7a}
\end{figure}

\begin{figure}[h]
    \centering
\includegraphics[width=0.5\linewidth]{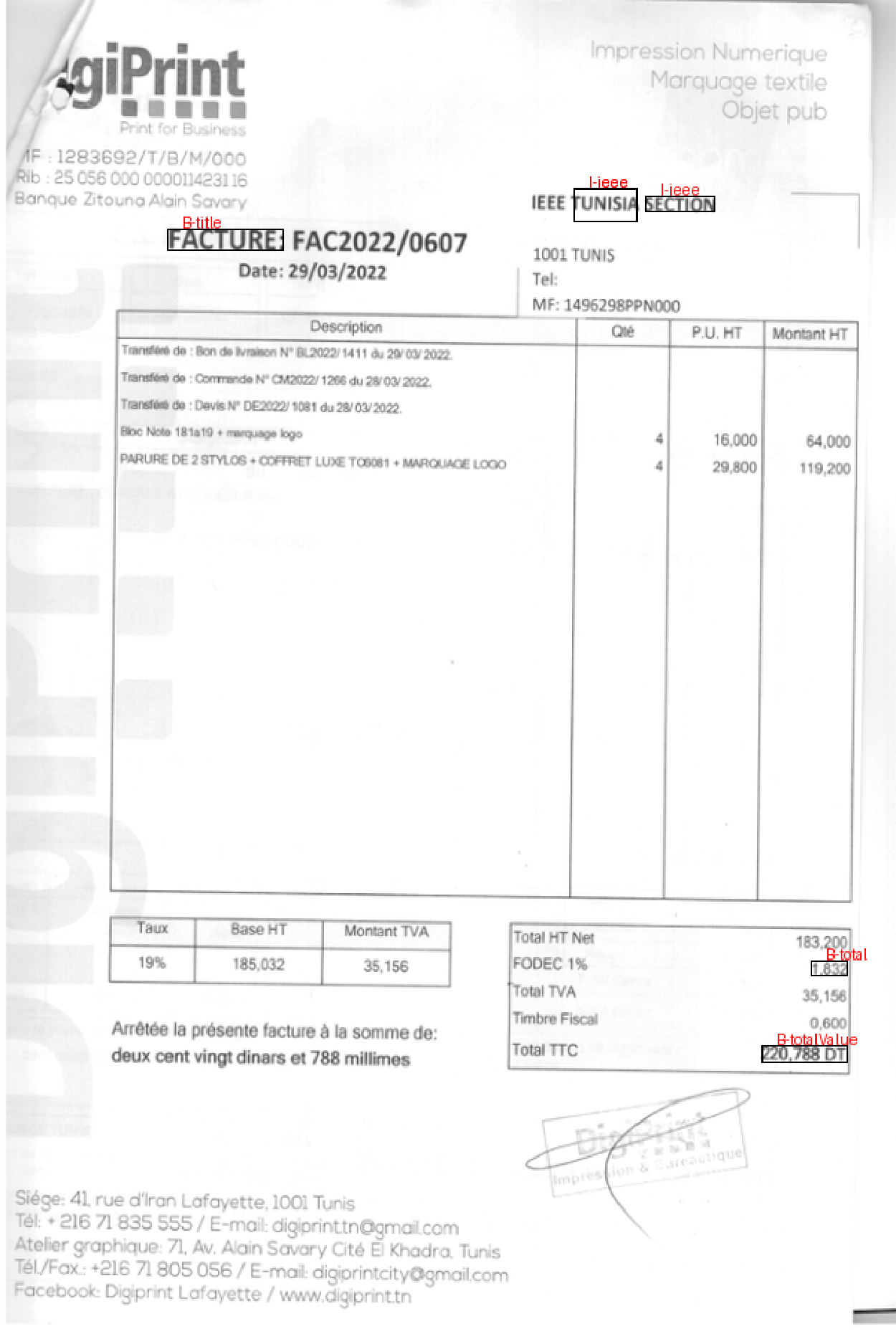}
    \caption{LayoutLMv3 output using Azure OCR.}
    \label{fig7b}
\end{figure}

We present the inference results with annotations powered by Tesseract (cf. Figure \ref{fig7a}) and Microsoft Azure OCR (cf. Figure \ref{fig7b}). As discussed earlier following the results of Table \ref{tab:class_imbalance}, Azure OCR provides more selective and stable results, capturing target fields with minimal errors, whereas Tesseract exhibits less precision, often mislabeling or capturing non-target fields.

In conclusion, the robustness and selectivity of Microsoft Azure OCR make it the preferred choice for the critical task of text recognition.

\subsection{Final keyword detection results}

For document layout analysis, we train models on 50 epochs with a batch size of 4. We opt for Adam optimizer and L2 regularization with a weight decay of $10^{-3}$. The learning rate is also fixed at $5e^{-5}$ for the three models.

\begin{table*}[h]
\centering
\caption{Comparison of document layout analysis models.}
\small

\begin{tabular}{lllllll}
\hline
\textbf{Model} & \textbf{GFlops} & \textbf{Inference Time (s)} & \textbf{Parameters} & \textbf{Validation Loss} & \textbf{F1-Score} & \textbf{Precision} \\
\hline
LayoutLMv3 & 44.83 & 2.08 & 125.93 millions & 3.6e-2 & \textbf{0.56} & \textbf{0.68} \\
\hline
LiLT & 43.2 & \textbf{1.04} &  130 millions & 3e-4 & 0.53 & 0.61 \\
\hline
LayoutLMv2 & 43.7 & 2.20 & 200 millions & 3.53e-2 & 0.47 & 0.49 \\
\hline
\end{tabular}
\label{tab:model-comparison-document-analysis}
\end{table*}

\begin{figure*}[h]
  \centering
  
  \subfloat[LayoutLMv3\label{fig:training-validation-loss_v3}]{\includegraphics[width=.3\linewidth]{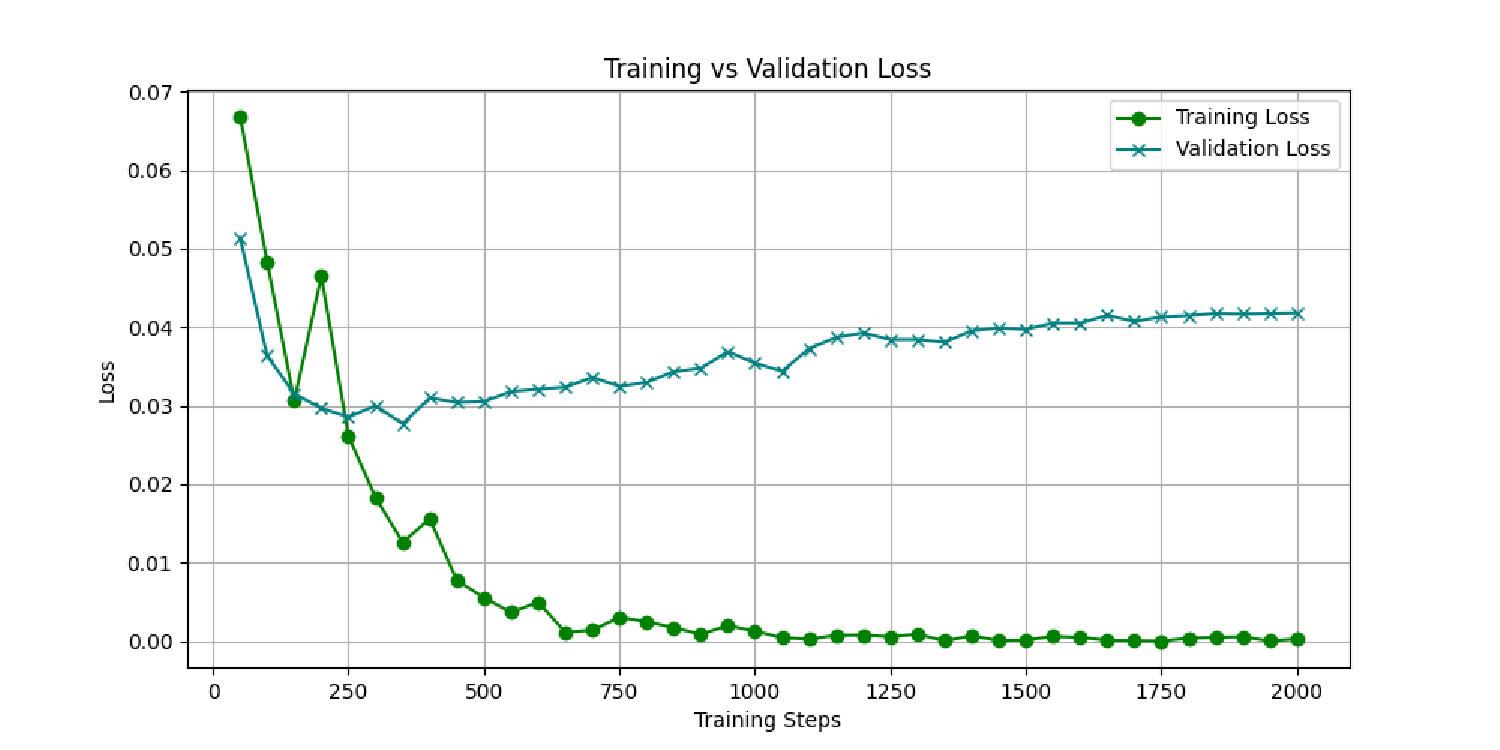}}
  \hfill
  \subfloat[LiLT\label{fig:training-validation-loss_lilt}] {\includegraphics[width=.3\linewidth]{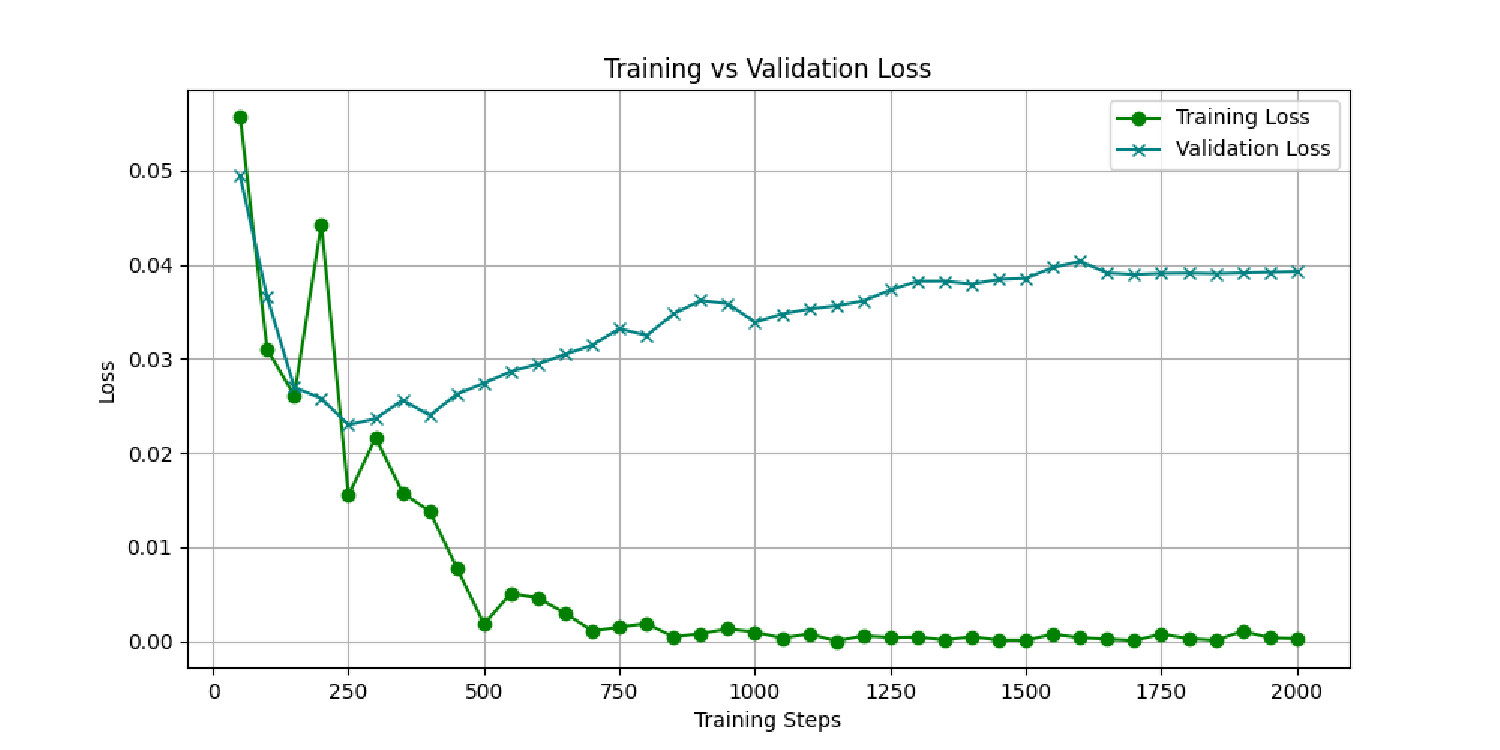}}
  \hfill
  \subfloat[LayoutLMv2\label{fig:training-validation-loss_v2}] {\includegraphics[width=.3\linewidth]{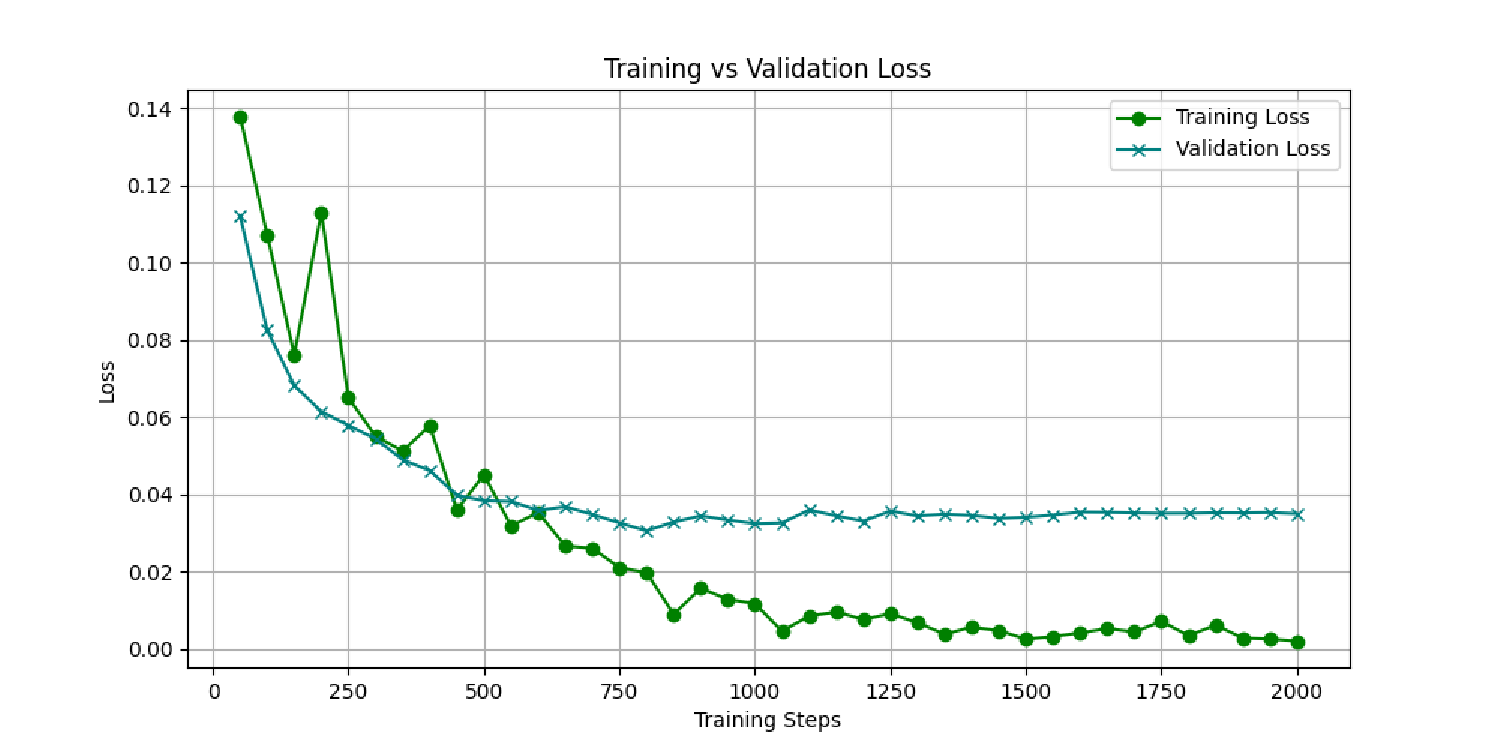}}
  
  \caption{Training and validation loss comparison for keyword detection.}
  \label{fig:training-validation-loss}%
\end{figure*}    

\begin{figure*}[h]
  \centering
  
  \subfloat[LayoutLMv3\label{fig:f1-score_v3}]{\includegraphics[width=.3\linewidth]{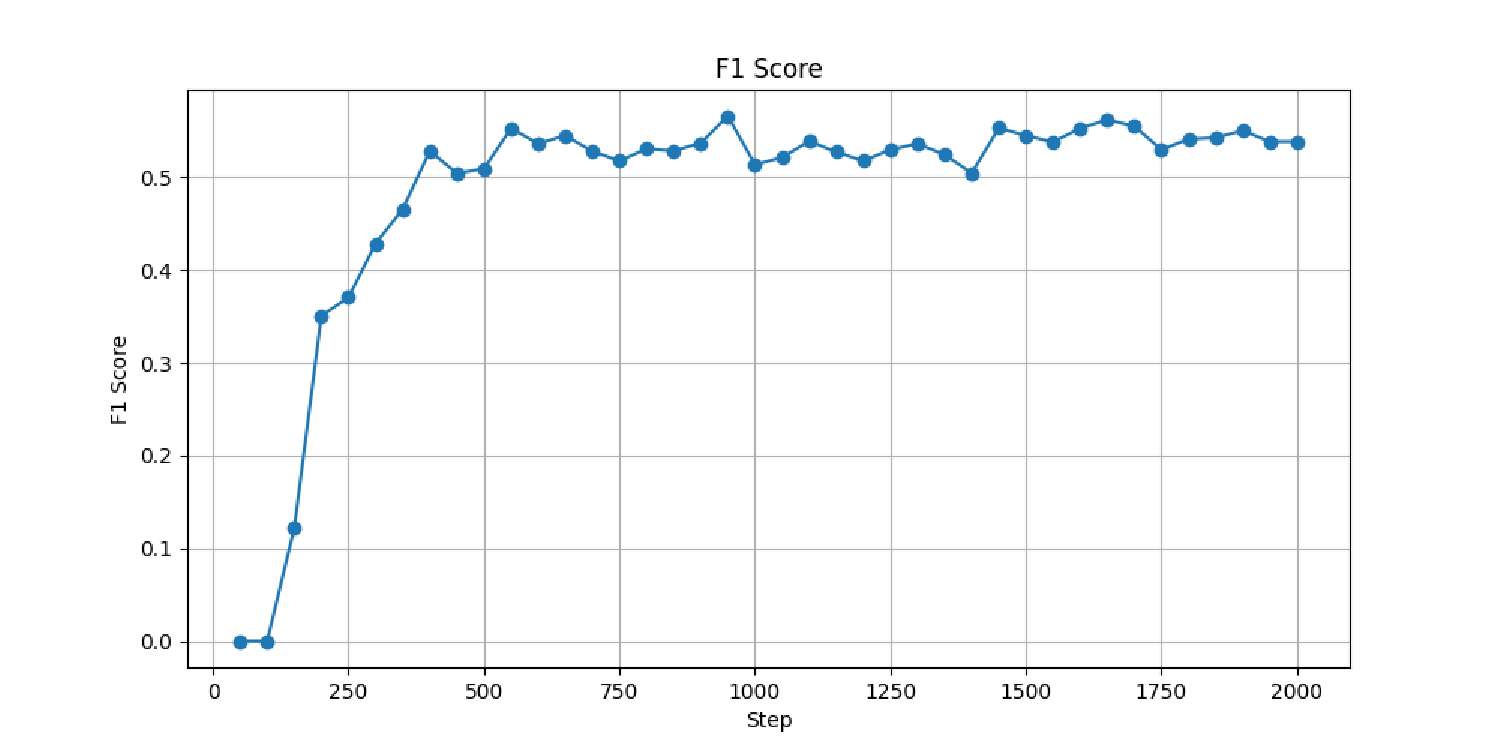}}
  \hfill
  \subfloat[LiLT\label{fig:f1-score_lilt}] {\includegraphics[width=.3\linewidth]{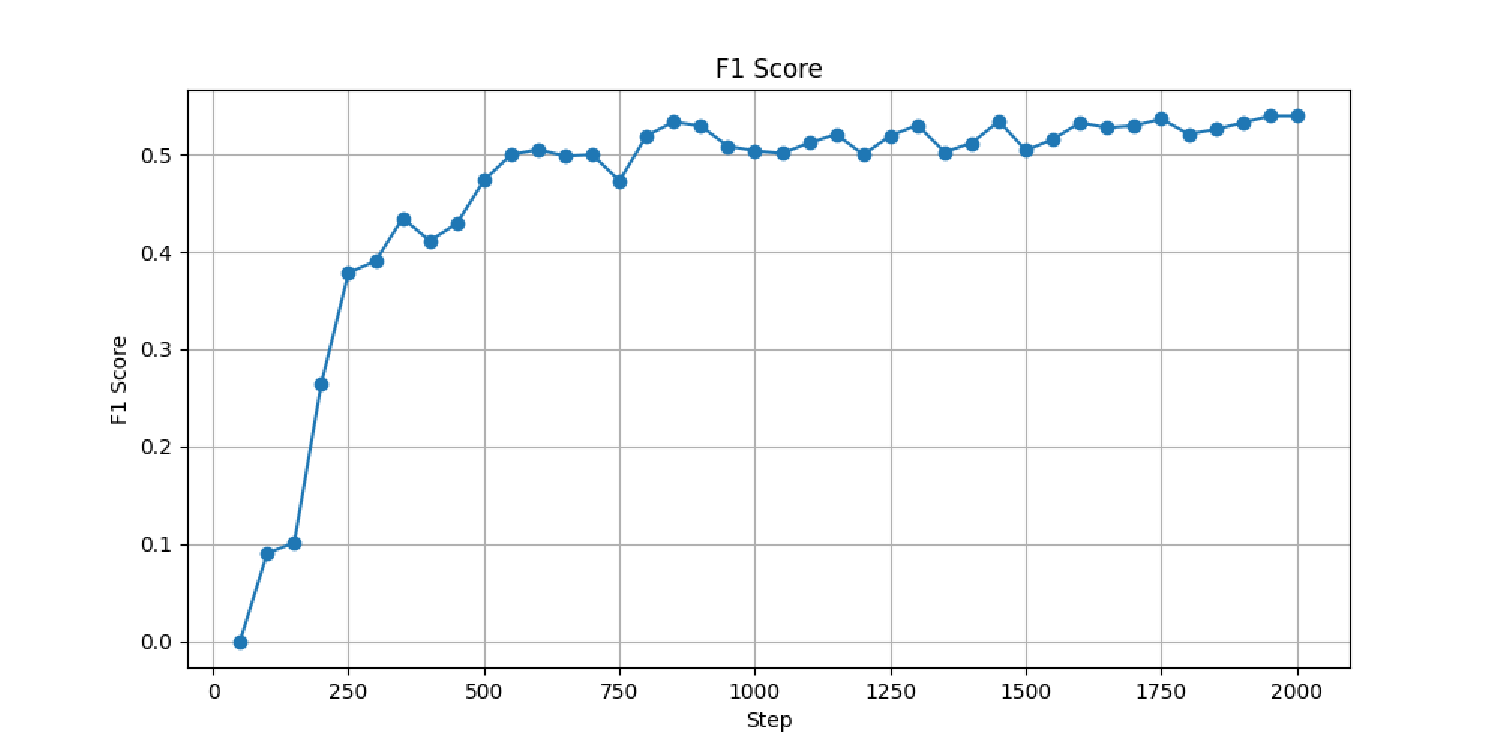}}
  \hfill
  \subfloat[LayoutLMv2\label{fig:f1-score_v2}] {\includegraphics[width=.3\linewidth]{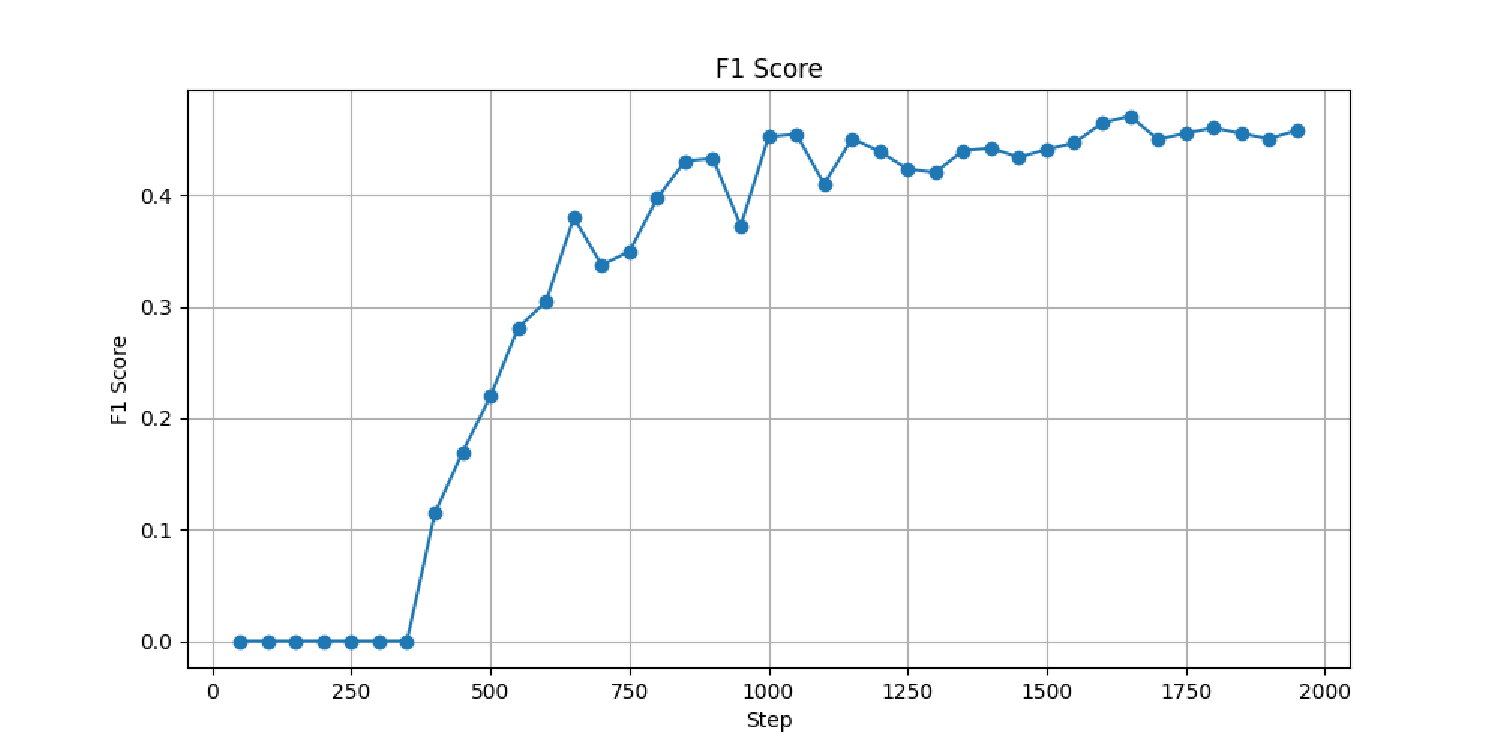}}
  
  \caption{F1-score over training steps comparison for keyword detection.}
  \label{fig:f1-score_lilt}%
\end{figure*} 

Addressing data imbalance is a crucial step in this process. We opt for using the focal loss, which is an altered version of weighted cross entropy designed to address class imbalance more effectively by focusing on hard-to-classify examples.

The remainder of this subsection presents the evaluation results of the three selected state-of-the-art document layout analysis models. Our analysis emphasizes both on the F1-score, for a balanced measure of precision and on recall, to highlight true positive rates. Table \ref{tab:model-comparison-document-analysis} summarizes the evaluation results and suggests that LayoutLMv3 performed the best, while LiLT required less inference time than the two other models.

Figure \ref{fig:training-validation-loss} shows the training and validation loss of the retained models. In all three plots, the training loss consistently decreases as the number of training steps increases. This suggests that the models are learning the training data effectively.  However, the ideal scenario is for the validation loss to also decrease alongside the training loss. However, we found that the validation loss for LayoutLMv3 (cf. Figure \ref{fig:training-validation-loss_v3}) starts to increase by around 1,250 steps, which is a clear indication of overfitting. LiLT (cf. Figure \ref{fig:training-validation-loss_lilt}) follows a similar but less severe pattern. LayoutLMv2 has the best performance, as depicted in Figure \ref{fig:training-validation-loss_v2}. The validation loss, after an initial slight decrease, can be seen to be fairly flat, thus indicating good generalization without overfitting the training data.

Based on these results, we conclude that overfitting was observed with both the LiLT and LayoutLMv3 models. However, they remain the top-performing models based on their least F1-scores. The overfitting issue could be mitigated by implementing an early stopping strategy. For interpretation purposes, we chose to continue training all models for the same number of epochs to show their overall tendencies. In future work, these models should be re-trained with early stopping, halting the training when no improvement in validation loss is observed after a certain number of steps.

Figure \ref{fig:f1-score_lilt} depicts the F1-scores over training for each model. All models show an increasing trend in F1-scores with more training steps, though the rate of improvement varies: LayoutLMv3 achieves the highest overall F1-score, indicating superior performance (Figure \ref{fig:f1-score_v3}), while both LiLT and LayoutLMv2 follow. Although not the fastest, LayoutLMv3's inference time remains acceptable (see Table \ref{tab:model-comparison-document-analysis}).

\subsection{Signature and stamp detection}
In this section, we present the results of training and validation of object detection models for signature and stamp detection. We mainly focus on training and validation loss, as well as mAP over epochs for each model. Results are summarized in Table \ref{tab:model-comparison}. All models were trained with a momentum of 0.9 and L2 regularization (weight decay of 0.0005). Key parameters for training include a batch size equal to 8, 50 epochs, and a learning rate of 0.001.

\begin{table*}[h]
\centering
\caption{Comparison of signature and stamp detection models.}
\small

\begin{tabular}{llllllll}
\hline
Model & \textbf{GFlops} & \textbf{Inference Time (s)} & \textbf{Parameters} & \textbf{Training Loss} & \textbf{Validation Loss} & \textbf{mAP@0.50} & \textbf{mAP@50:95} \\
\hline
RetinaNet       & 128.32 & 0.29 & 36.4 millions & 0.22 & 0.38 & \textbf{83.96} & \textbf{45.70} \\ \hline
Faster R-CNN    & 134.38 & 0.18 & 41 millions & 0.11 & 0.20 & 81.351 & 42.085 \\ \hline
FCOS            & 80.34 & 0.17 & 32.1 millions & 0.73 & 1.18 & 76.94 & 41.77 \\ \hline
SSD             & 30.48 & \textbf{0.11} & 23.9 millions & 1.33 & 3.59 & 68.33 & 31.19 \\
\hline
\end{tabular}
\label{tab:model-comparison}
\end{table*}

\begin{figure*}[h]
  \centering
  \subfloat[RetinaNet\label{fig:training_validation_loss_retinanet}]{\includegraphics[width=.215\linewidth]{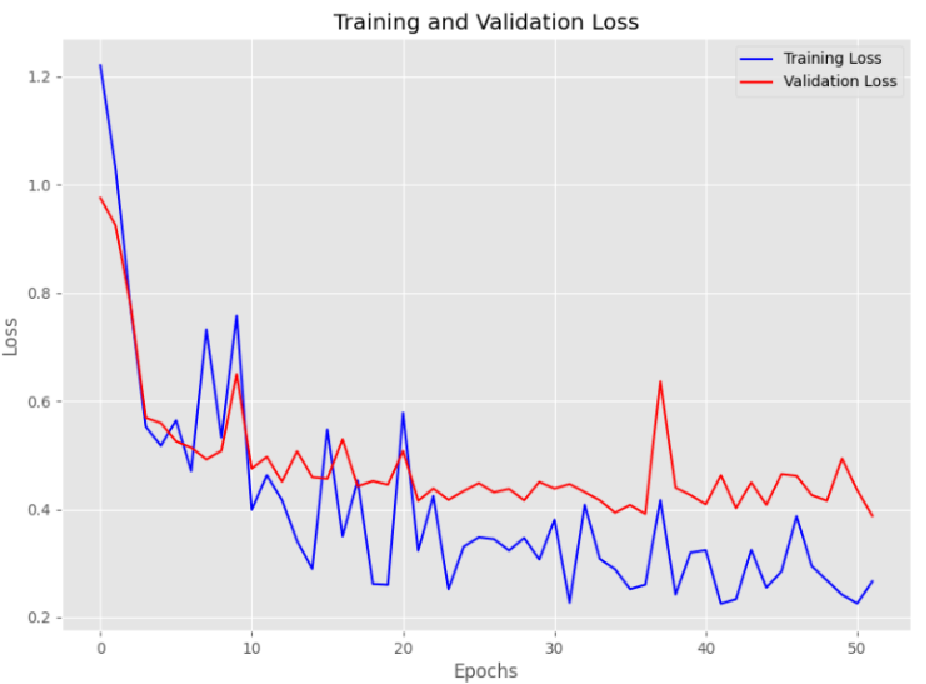}}
  \hfill
  \subfloat[Faster R-CNN\label{fig:training_validation_loss_fasterrcnn}] {\includegraphics[width=.25\linewidth]{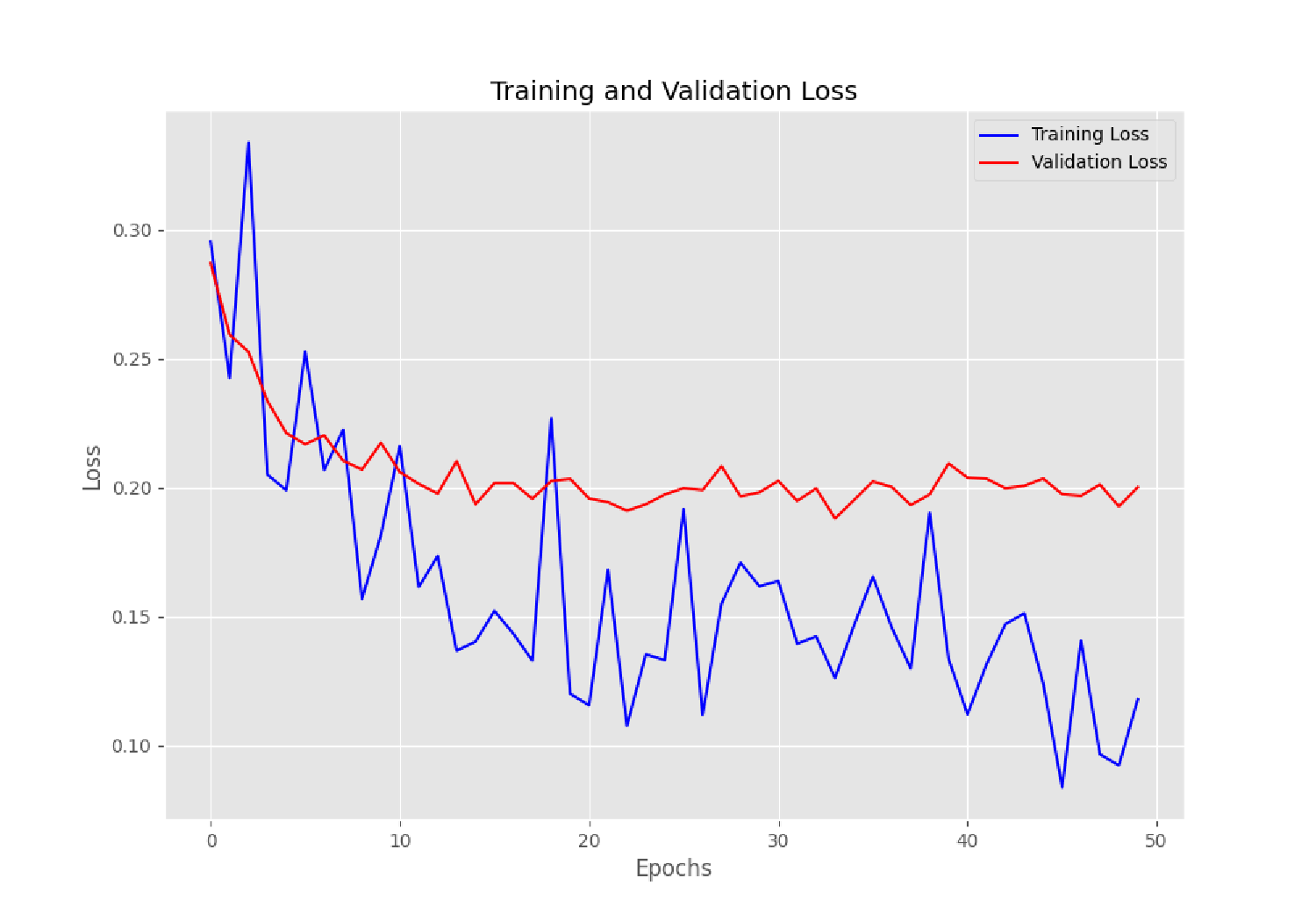}}
  \hfill
\subfloat[FCOS\label{fig:training_validation_loss_fcos}] {\includegraphics[width=.25\linewidth]{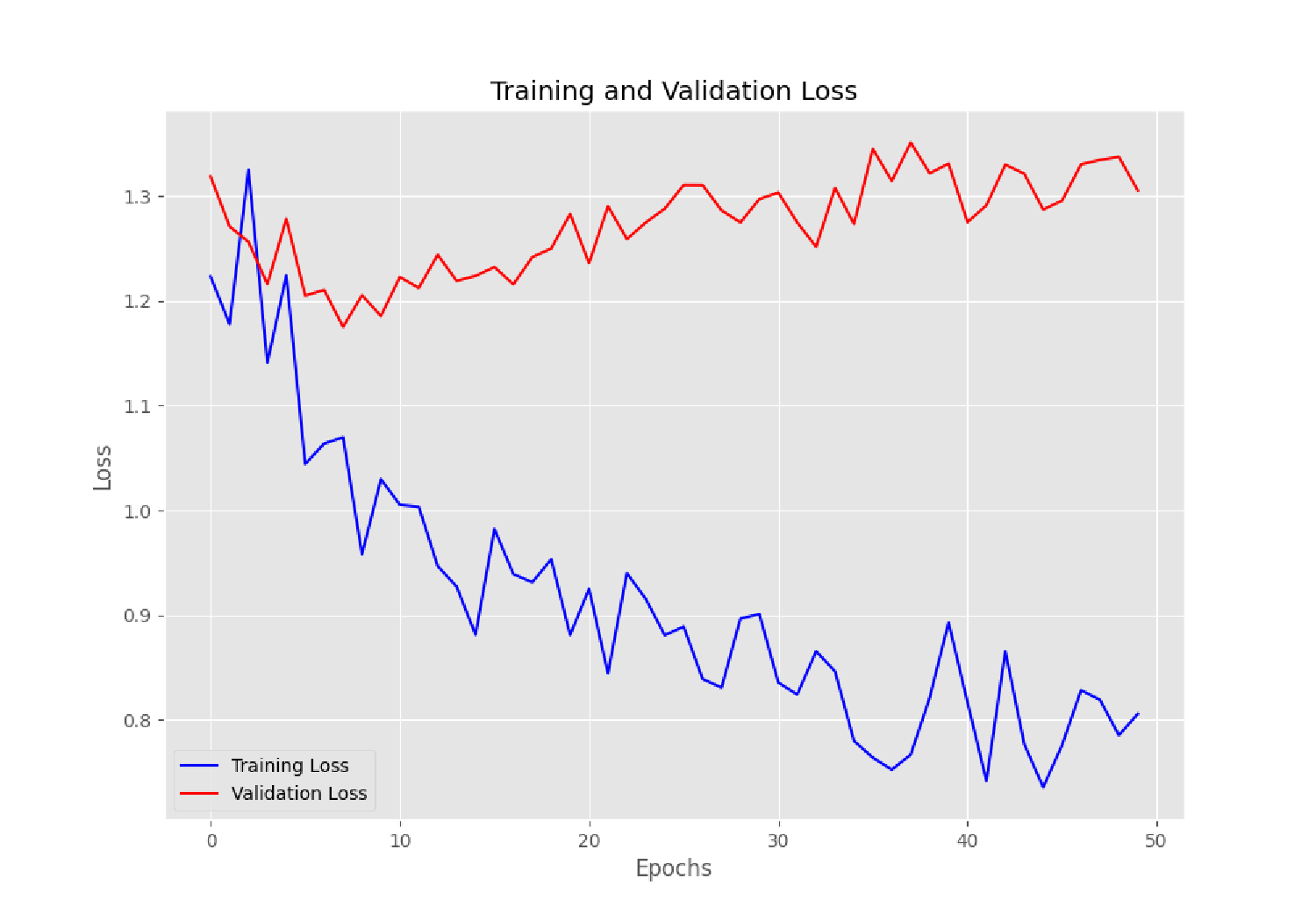}}
  \hfill
  \subfloat[SSD\label{fig:training_validation_loss_ssd}] {\includegraphics[width=.25\linewidth]{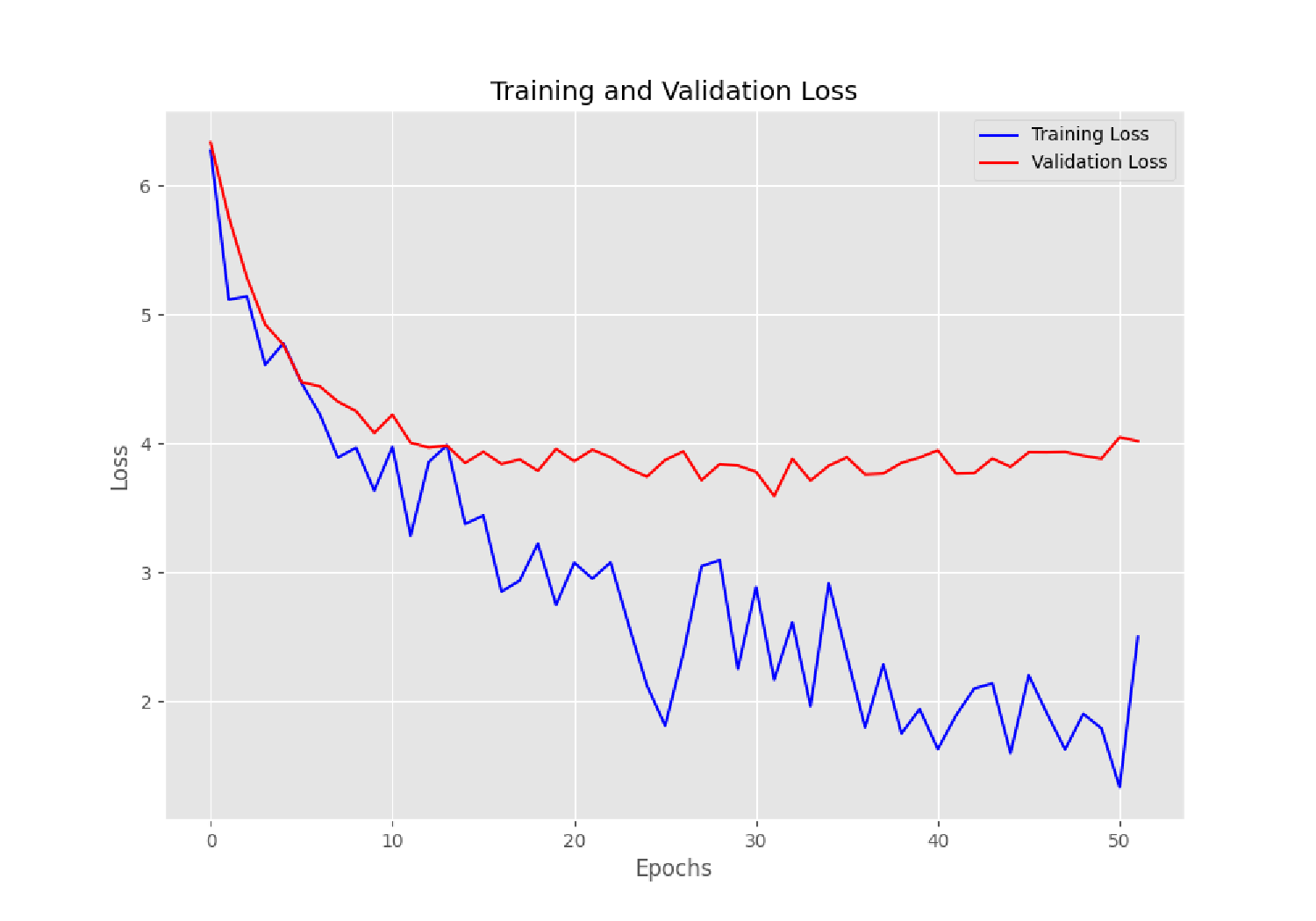}}
  \caption{Training and validation loss comparison for signature and stamp detection.}
  \label{fig:training_validation_loss_comparison}
\end{figure*}

\begin{figure*}[h]
  \centering
  \subfloat[RetinaNet\label{fig:map_retinanet}]{\includegraphics[width=.215\linewidth]{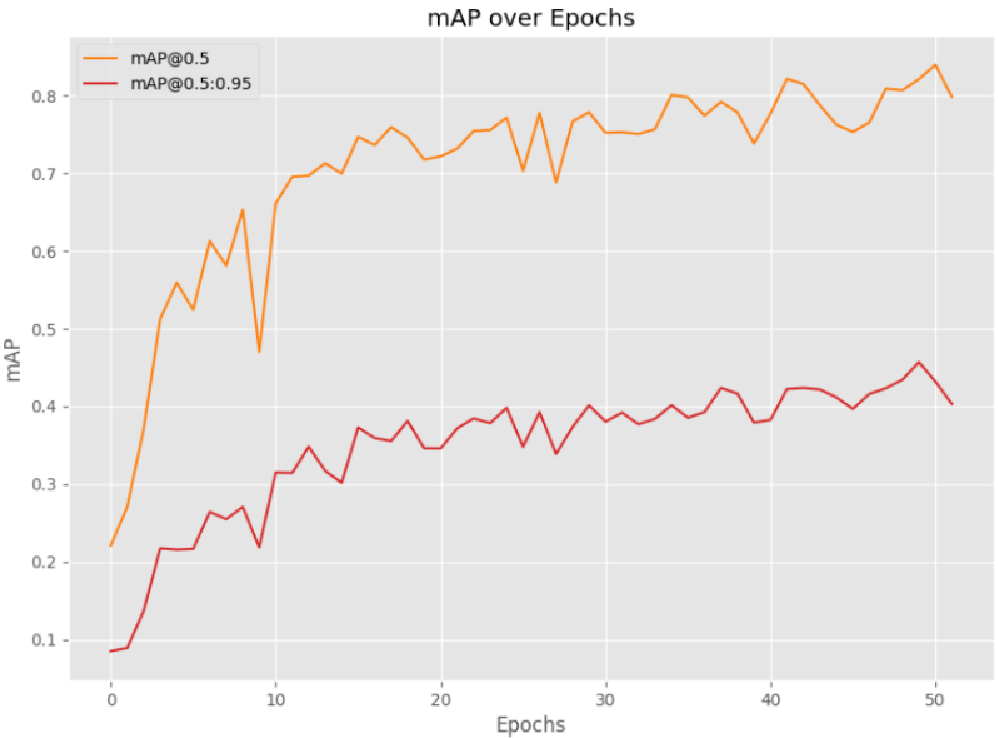}}
  \hfill
  \subfloat[Faster R-CNN\label{fig:map_fasterrcnn}] {\includegraphics[width=.25\linewidth]{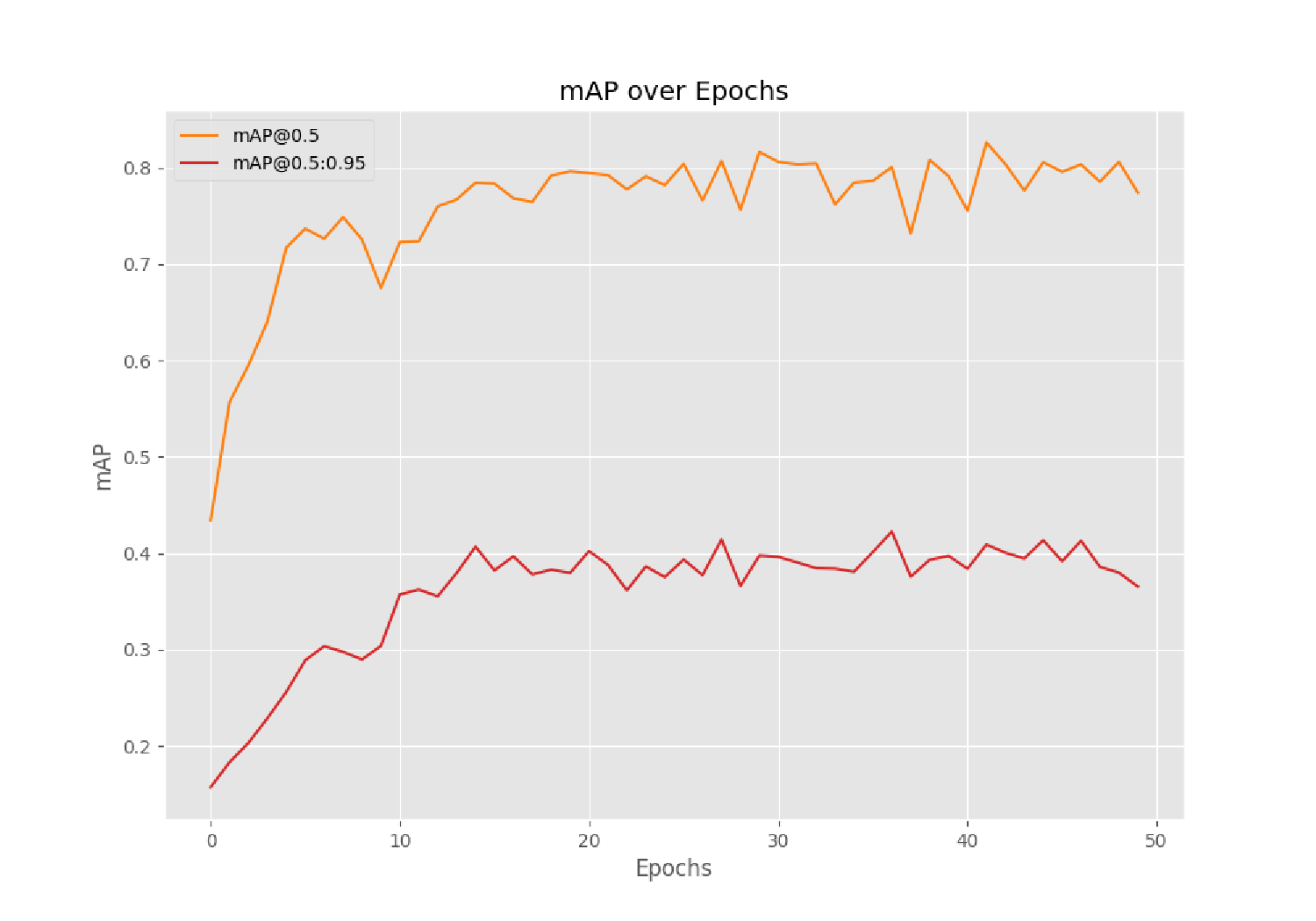}}
  \hfill
  \subfloat[FCOS\label{fig:map_fcos}] {\includegraphics[width=.25\linewidth]{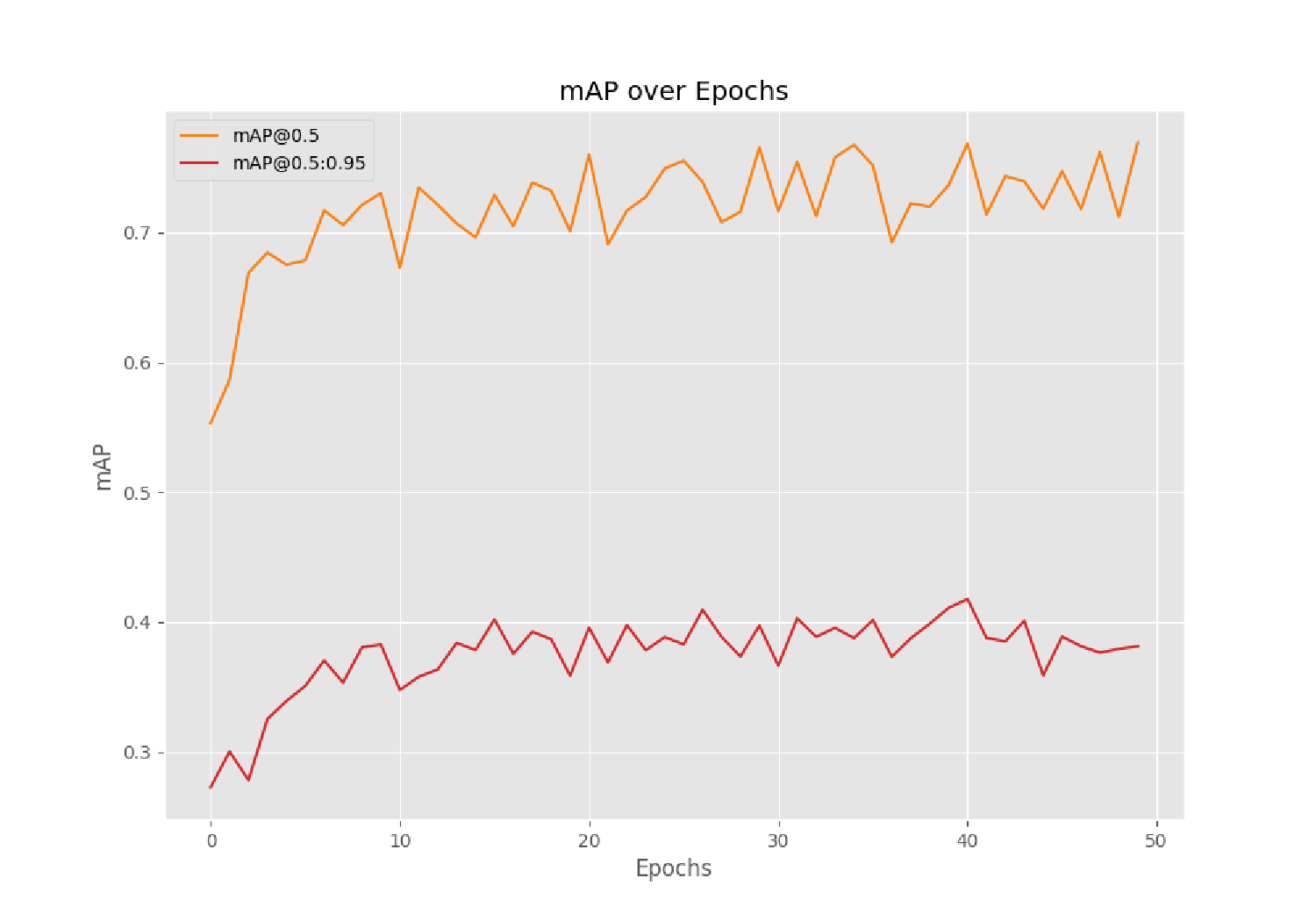}}
  \hfill
  \subfloat[SSD\label{fig:map_ssd}] {\includegraphics[width=.25\linewidth]{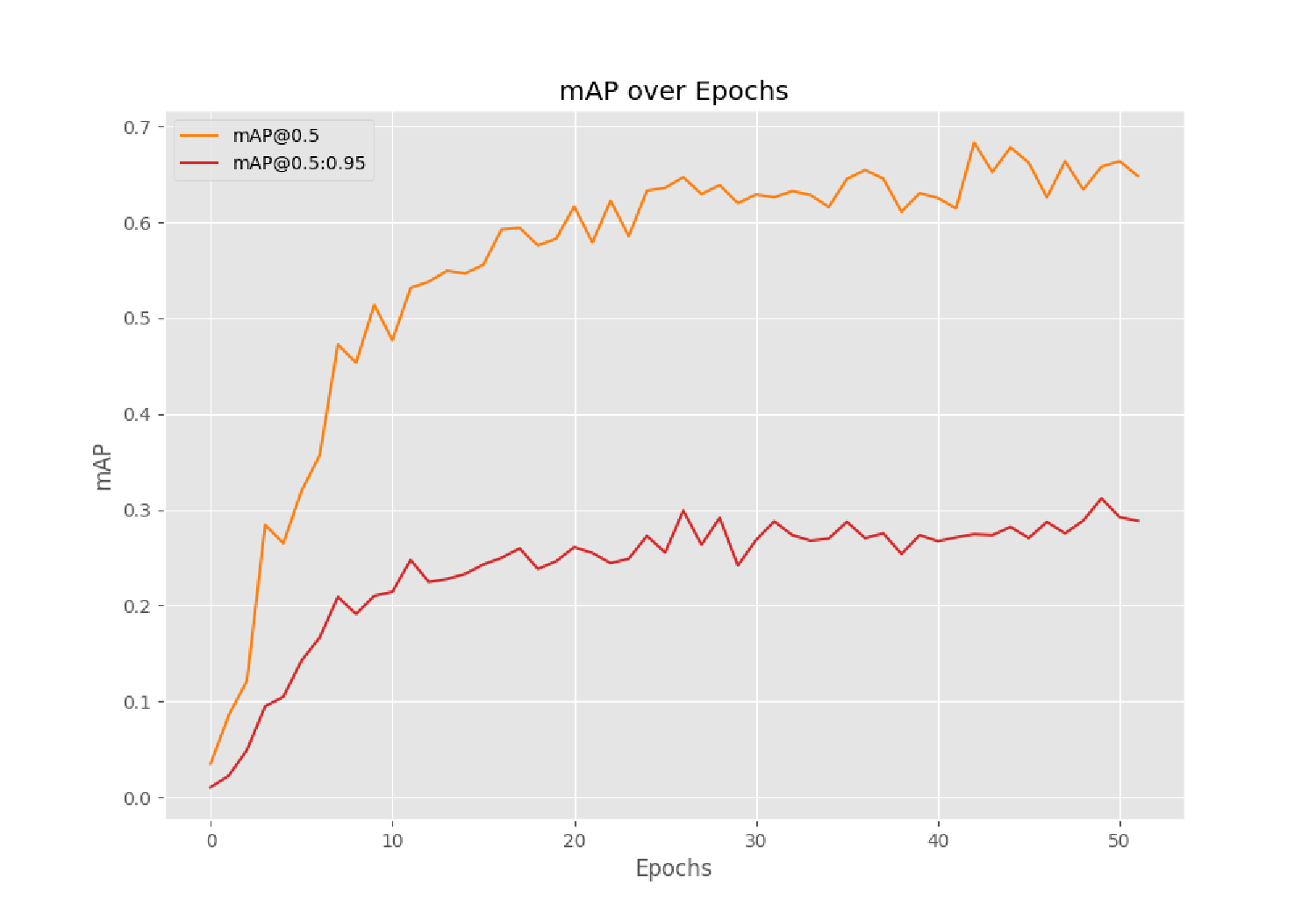}}
  \caption{mAP comparison for signature and stamp detection}
  \label{fig:map_comparison}
\end{figure*}

Examining the training and validation loss is important for evaluating the performance of object detection models. Based on the results shown in Figure \ref{fig:training_validation_loss_comparison}, we note that RetinaNet (Figure \ref{fig:training_validation_loss_retinanet}) shows a significant decrease in training loss, indicating effective learning, while validation loss initially decreases but later fluctuates, indicating potential overfitting. Faster R-CNN (Figure \ref{fig:training_validation_loss_fasterrcnn}) follows a similar trend. Meanwhile, both FCOS (Figure \ref{fig:training_validation_loss_fcos}) and SSD (Figure \ref{fig:training_validation_loss_ssd}) have decreasing training losses with relatively high validation losses.

We conclude that RetinaNet demonstrates superior generalization ability, Faster R-CNN remains competitive with some overfitting, while FCOS and SSD suffer from significant overfitting, impacting their overall performance and generalization.

Figure \ref{fig:map_comparison} depicts the mAP over epochs for the four considered models. We note that RetinaNet (Figure \ref{fig:map_retinanet}) quickly stabilizes around 0.84 mAP@0.50, indicating high detection accuracy, with a steady increase in mAP@50:95. Faster R-CNN (Figure \ref{fig:map_fasterrcnn}) achieves 0.81 mAP@0.50, slightly below RetinaNet, and maintains consistent performance in mAP@50:95. FCOS (Figure \ref{fig:map_fcos}) reaches 0.77 mAP@0.50 but shows lower and more variable mAP@50:95 compared to previous models. Finally, SSD (Figure \ref{fig:map_ssd}) has the lowest 0.68 mAP@0.50 and variability in mAP@50:95 highlighting challenges in achieving good precision. 

\begin{figure}[h]
        \centering
        \includegraphics[width=0.48\textwidth]{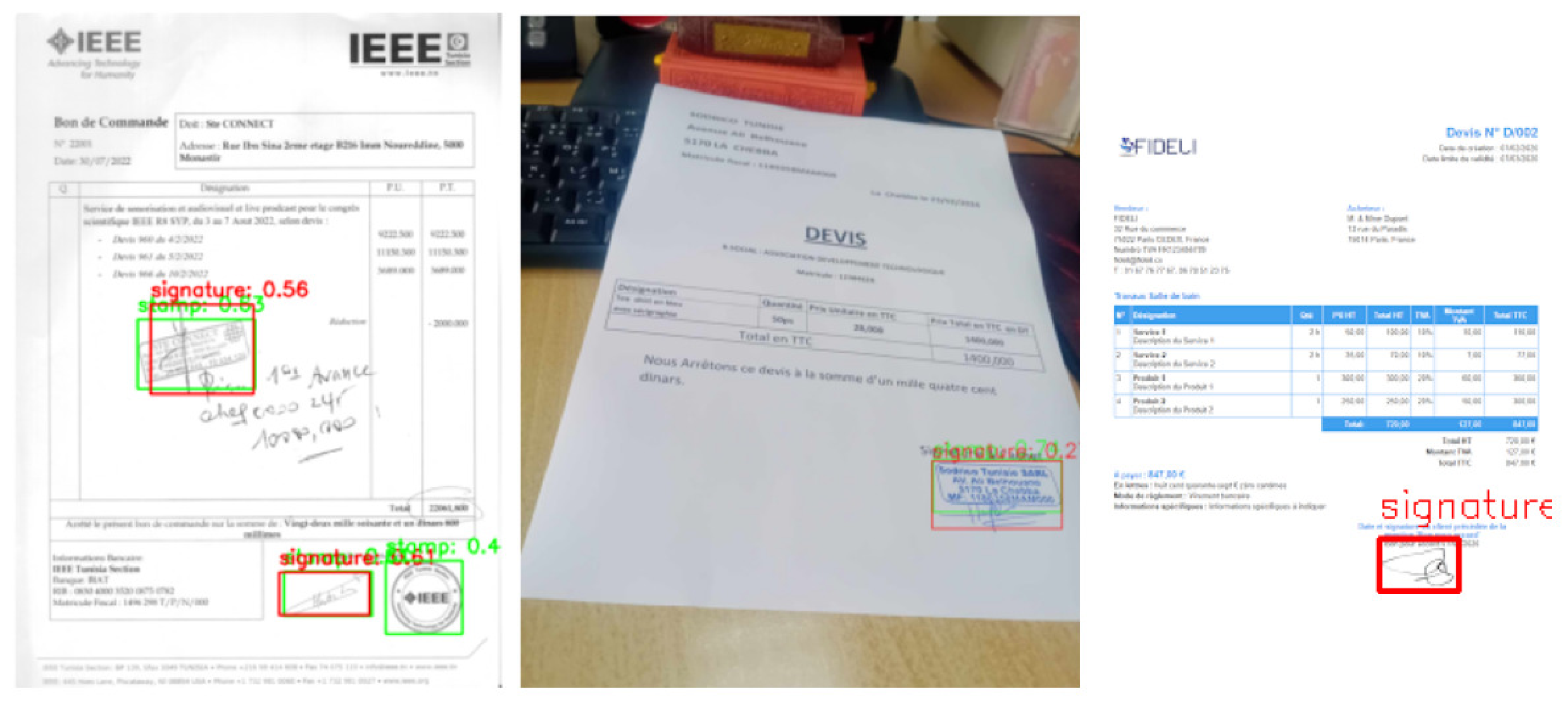}
        \caption{Inference results for RetinaNet.}
        \label{fig:inference_results_retinanet}
    \end{figure}

In summary, RetinaNet has superior detection accuracy and stability across IoU thresholds, while Faster R-CNN and FCOS perform well but lag behind in accuracy under stricter IoU conditions. SSD faces significant challenges in achieving high precision across all thresholds. Given these results, RetinaNet stands out as the best model. It offers the highest detection accuracy with reasonable computational efficiency, although not the fastest. The inference results for RetinaNet (see Figure \ref{fig:inference_results_retinanet}) demonstrate its robustness in accurately detecting and localizing signatures and stamps across various document types, further reinforcing its suitability for the signature and stamp detection task.\\

Based on the comparative analysis of the different models, and to guarantee a robust and efficient pipeline, we select the following methods:
\begin{itemize}
    \item For keyword recognition, we move forward with Microsoft Azure OCR for the OCR step, and LayoutLMv3 for document layout analysis step. 
    \item For signature and stamp detection, we move forward with RetinaNet.
\end{itemize}

\section{Conclusion\label{sec:concl}}
To address the urgent need for efficient solutions to automate the validation of invoice documents, we propose a novel, fully annotated dataset composed of real invoices with various imperfections, along with a robust approach that accounts for different document layouts and real-world data variations. This automation not only streamlines the validation process but also significantly reduces human error and enhances operational efficiency. Our contribution is twofold: first, we focus on keyword detection using OCR and document layout analysis models; second, we identify stamps and signatures using object detection models. For each task, we benchmark various DL models to establish a strong foundation that balances efficiency and performance. Based on our evaluation results, we select an end-to-end system combining Microsoft Azure OCR with LayoutLMv3 for the keyword detection task, while RetinaNet proves to be the best choice for stamp and signature detection. Future work will involve further hyperparameter tuning and refinement of the DL models. Additionally, we plan to focus on handwritten invoices, which pose unique challenges due to their high variability in styles and legibility issues.

\section*{Acknowledgments}
This work is the result of a collaboration between the IEEE Tunisia Section Volunteers Tools Coordination and the National Institute of Applied Sciences and Technology (INSAT). The authors would like to acknowledge all the members of the IEEE Tunisia Section Units Accounting Platform development team for their valuable efforts.

\end{document}